\documentclass[10pt,twocolumn,letterpaper]{article}

\usepackage[pagenumbers]{cvpr} 

\definecolor{cvprblue}{rgb}{0.21,0.49,0.74}
\usepackage[pagebackref,breaklinks,colorlinks,allcolors=cvprblue]{hyperref}

\def\paperID{xxxx} 
\def\confName{CVPR}
\def\confYear{2026}

\title{TF-Lane: Traffic Flow Module for Robust Lane Perception}

\author{
  Yihan Xie\textsuperscript{†, *}, \enspace Han Xia\textsuperscript{†}, \enspace Zhen Yang \\
  BYD Company Limited \\
  \texttt{\small aerocooc@gmail.com}
}

\begin{document}
\maketitle
\footnotetext[1]{† Equal contribution. * Corresponding author.}
\begin{abstract}
Autonomous driving systems require robust lane perception capabilities, yet existing vision-based detection methods suffer significant performance degradation when visual sensors provide insufficient cues, such as in occluded or lane-missing scenarios. While some approaches incorporate high-definition maps as supplementary information, these solutions face challenges of high subscription costs and limited real-time performance. To address these limitations, we explore an innovative information source: traffic flow, which offers real-time capabilities without additional costs. This paper proposes a TrafficFlow-aware Lane perception Module (TFM) that effectively extracts real-time traffic flow features and seamlessly integrates them with existing lane perception algorithms. This solution originated from real-world autonomous driving conditions and was subsequently validated on open-source algorithms and datasets. Extensive experiments on four mainstream models and two public datasets (Nuscenes and OpenLaneV2) using standard evaluation metrics show that TFM consistently improves performance, achieving up to $+4.1\%$ mAP gain on the Nuscenes dataset.
\end{abstract}    
\section{Introduction}
\label{sec:intro}

\begin{figure}[h]
  \centering
   \includegraphics[width=1\linewidth]{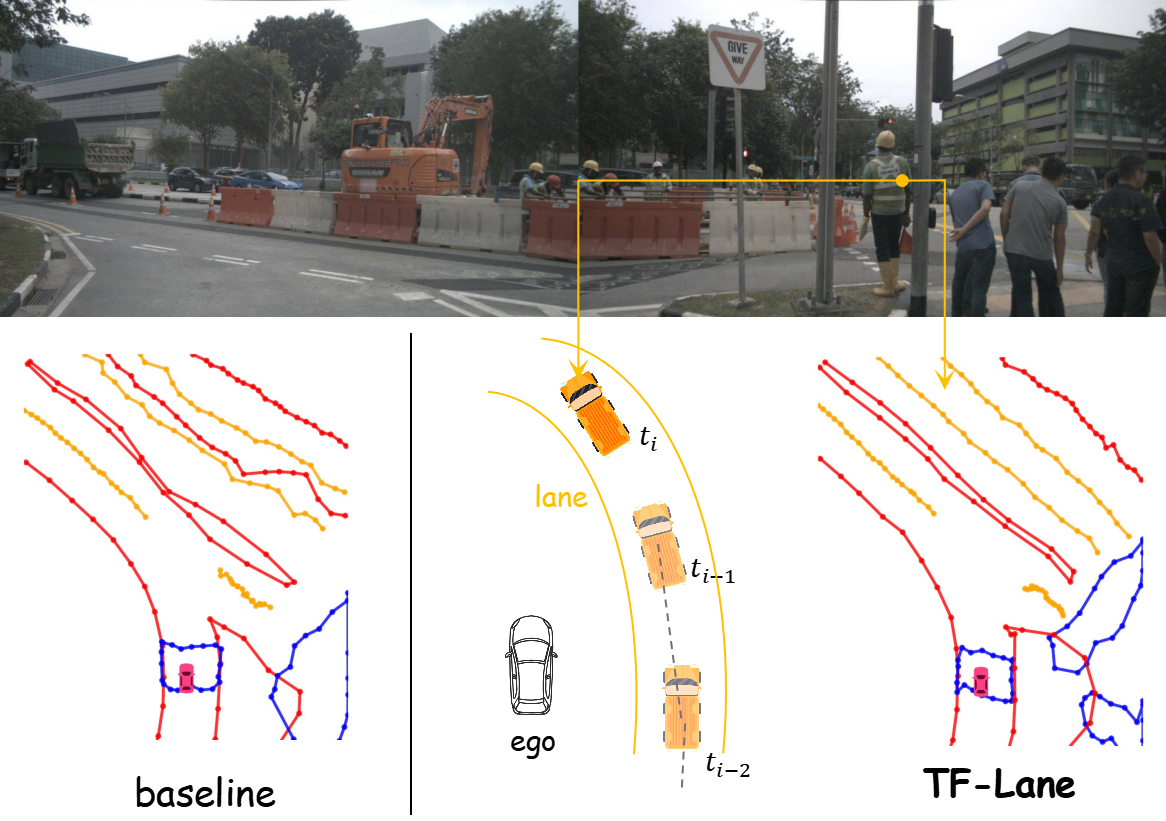}
   \caption{\textbf{Illustration of Traffic Flow.} In occluded or complex scenes, traffic flow information enhances the robustness of lane perception by providing real-time prior info.}
   \label{fig:onecol}
   \vspace{-0.1cm}
\end{figure}

The rapid advancement of autonomous driving technology underscores its critical role in intelligent transportation systems. As a core component of these systems, robust lane detection is essential. Existing approaches for lane perception often rely on monocular or multi-view vision-based models. However, in complex scenarios such as traffic congestion or road construction, visual sensors alone may fail to capture sufficient cues, leading to performance degradation. To address this, prior work often incorporates prior knowledge as auxiliary information, with high-definition (HD) maps being a mainstream solution. Yet, HD maps face limitations including high subscription costs and poor real-time performance. In this paper, we observe that autonomous driving, as an integrated task, can leverage another prior information source for road structure perception: traffic flow, which offers real-time capabilities without additional costs.

In autonomous driving systems, traffic flow serves as a real-time prior information source, comprising continuous streams of vehicles and broadly encompassing other traffic participants (e.g., pedestrians). The state of traffic flow dynamically reflects road structure in real-time, as it inherently implies that regions traversed by preceding vehicles within a limited time are navigable for the ego-vehicle, whereas areas with dense pedestrian presence are non-navigable in the short term. This information can be efficiently acquired via internal system nodes subscribing to dynamic object detection and tracking modules, offering high real-time performance and low communication costs without additional financial investment. Consequently, traffic flow provides a robust and economical auxiliary prior for lane detection, particularly in complex scenarios where visual cues are insufficient.

\begin{figure*}[t]
  \centering
  \includegraphics[width=1\textwidth]{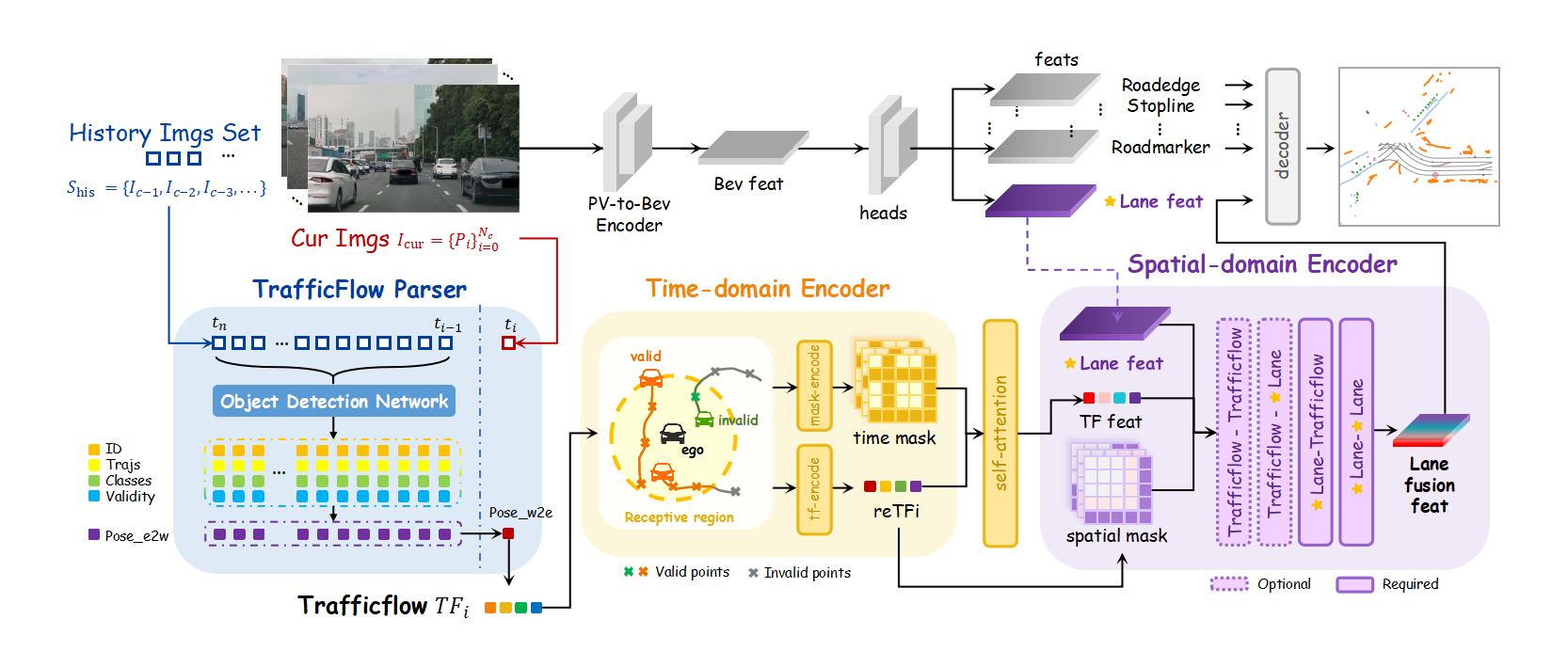}
  \caption{\textbf{TF-Lane Overall Architecture.} The proposed TFM module consists of three key components: Traffic Flow Extraction, Temporal Feature Encoding, and Spatial Feature Fusion. The fused lane features are then processed by the decoder to obtain the final lane perception results in the scene.}
  \label{fig:netarch}
  \vspace{-0.3cm}
\end{figure*}

To enhance the robustness of lane detection under challenging scenarios (e.g., occlusions or missing lanes), we propose TF-Lane, a framework that integrates a Traffic Flow feature fusion Module (TFM) as its core component. As illustrated in \cref{fig:netarch}, TFM takes traffic flow data as input and first processes it through a trajectory parser to encode raw information into traffic flow features, including object categories and spatial locations. A region sampling strategy is then applied to concentrate on scene-relevant features, ensuring efficient and focused utilization of traffic flow cues.

To address the challenges of occlusions and sensor instability in real-world autonomous driving, we propose a spatio-temporal dual-encoder architecture. The temporal encoder constructs corresponding masks and processes trajectory associations across consecutive frames, along with compensation for frame discontinuities, to capture the dynamic evolution of traffic flow through temporal modeling. The spatial encoder establishes a fusion mapping between traffic flow instances and lane features, integrating contextual traffic flow information with visual cues to enhance the robustness of lane representation in challenging scenarios.

To validate the efficacy of the Traffic Flow Module (TFM), we integrate TFM into four mainstream lane detection models: TopoNet\cite{ben2022toponet}, LaneSegNet\cite{li2023lanesegnet}, Maptr\cite{liao2022maptr}, Maptrv2\cite{liao2025maptrv2}, and evaluate them on the NuScenes\cite{caesar2020nuscenes} and OpenLaneV2\cite{wang2023openlane} datasets. We also introduce a traffic flow dataset construction method that generalizes across multiple open-source datasets, complemented by real-world vehicle validation to ensure broad applicability. For fairness, we adhere to each model's original evaluation metrics. As depicted in \cref{fig:multi-model-infer}, TFM integration yields consistent performance gains across all models. Ablation studies further substantiate the contributions of individual components within TFM.

Notably, while the TF-Lane framework supports explicit traffic flow input during both training and inference, experiments demonstrate that even when traffic flow is omitted at inference, the model outperforms the baseline. This underscores that traffic flow imparts effective implicit supervision during training, bolstering the robustness of lane representations.

The primary contributions are summarized as follows:

\begin{itemize}
\item This paper pioneers the integration of traffic flow information into lane perception tasks, introducing an innovative prior that leverages high real-time capability and zero cost, improving robustness of visual-based methods in challenging scenarios such as occlusions or missing lanes.
\item We propose TF-Lane, a framework that incorporates traffic flow and visual features through a Traffic Flow-aware Module (TFM). The framework employs a spatio-temporal dual-encoder architecture to efficiently extract and fuse features, enabling flexible adaptation to diverse lane detection tasks. It significantly enhances perception robustness in complex environments and achieves consistent quantitative improvements across multiple benchmarks.
\end{itemize}

\section{Related Work}
\label{sec:formatting}

\subsection{Vision-based Approaches}

Early lane detection methods primarily followed segmentation-based paradigms, formulating the task as either semantic segmentation \cite{parashar2017scnn} or instance-level segmentation \cite{qiu2022priorlane}. For example, SCNN treats each lane as an independent category for pixel-wise classification, while Prior-Lane introduces structural priors to enhance geometric consistency, laying the groundwork for subsequent map-prior fusion methods. However, these approaches require predefining the maximum number of lanes, limiting their flexibility in dynamic scenes.

To overcome these constraints, research shifted to bottom-up instance segmentation strategies. LaneNet \cite{wang2023bev} and LaneAF\cite{abualsaud2021laneaf} achieve instance discrimination without preset counts by clustering binary lane-background masks. With the growing demand for 3D perception in autonomous driving, lane detection has expanded into 3D space. 3D-LaneNet\cite{garnett20193d} first proposed an end-to-end 3D lane detection framework, and subsequent works like Gen-LaneNet\cite{guo2020gen} and 3D-LaneNet+\cite{efrat20203d} project image features into bird's-eye view (BEV) space via inverse perspective mapping (IPM). To mitigate IPM's geometric prior errors, recent studies adopt learnable view transformation modules (e.g., LSS\cite{philion2020lift}, BevDet\cite{huang2021bevdet}, and BevFormer\cite{li2024bevformer}) for more robust BEV feature construction.

Concurrently, lane structure modeling has emerged as another key direction. TopoNet\cite{ben2022toponet} pioneers end-to-end lane topology reasoning, LaneSegNet\cite{li2023lanesegnet} jointly optimizes geometric structures and semantic elements, and Transformer-based MapTR series\cite{liao2022maptr,liao2025maptrv2} directly output vectorized lane representations through query mechanisms, significantly improving the efficiency and accuracy of structured outputs. Despite the strong performance of these pure vision methods in ideal conditions, their perceptual robustness remains challenged in complex environments with limited sensor cues, such as heavy occlusion or extreme illumination.

\subsection{Prior-based Approaches}

Perceptual uncertainty of onboard sensors in complex environments has motivated the integration of prior knowledge to enhance detection robustness. Map priors, owing to their inherent capacity for representing road structures, have emerged as a predominant choice.

In explicit map encoding, MapLite2.0~\cite{ort2022maplite} constructs local high-definition (HD) map representations by combining Standard Definition (SD) maps with online sensor data. PriorLane~\cite{qiu2022priorlane} encodes SD maps into binary images and aligns them with the ego-coordinate system via a Keypoint Alignment Module (KEA) for fusion with visual features. SMERF~\cite{duckworth2024smerf} extracts ground elements from SD maps, generates structured features through a dedicated map encoder, and enhances road structure perception by interacting BEV features with map cues using a Map Cross-Attention mechanism. Similarly, MapEX~\cite{ho2025mapex} and P-MapNet~\cite{jiang2024p} achieve implicit fusion of map priors through learnable query vectors. To address localization errors in map matching, U-BEV~\cite{camiletto2024u} introduces a template matching mechanism that improves fusion accuracy by identifying optimal matching points.

However, inherent drawbacks of HD maps, such as high subscription costs and update latency, have driven exploration into alternative priors. Autograph~\cite{zurn2023autograph} pioneers the use of vehicle trajectory data for inference but relies on offline data construction from bird's-eye views. Jia~et~al.~\cite{jia2025enhancing} reduce HD map dependency by fusing crowd-sourced trajectories with map priors, yet still incur additional data subscription costs and limited real-time performance. These solutions continue to face dual challenges of cost and timeliness in practical onboard deployment.

\section{Method}

To address the limitations of existing approaches, we propose TF-Lane, a framework that effectively integrates traffic flow with visual features via a Traffic Flow-aware Module (TFM), offering the advantages of zero additional cost and high real-time performance. This section elaborates on the overall framework, data preparation, and model architecture.

\subsection{Overall Framework}

As illustrated in \cref{fig:netarch}, the TF-Lane framework comprises three core components:

\paragraph{Traffic Flow Extraction:} Taking historical $n$ frames of sensor data as input, we first obtain attributes of traffic participants using an object detection algorithm. The sequential historical information is then transformed into the current frame's coordinate system based on localization data, resulting in the raw traffic flow representation $TF_i$. This process leverages inherent communication nodes of the onboard system without requiring additional data subscriptions.

\vspace{-0.3cm}
\paragraph{Temporal Feature Encoding:} The $TF_i$ is fed into a temporal encoder to extract effective features within the perceptual range of the ego vehicle. A designed masking mechanism filters out invalid frames caused by tracking interruptions, introducing refined traffic flow features $reTF_i$ and a temporal mask. This module employs self-attention to enhance the representation of features and generates a corresponding spatial mask.

\vspace{-0.3cm}
\paragraph{Spatial Feature Fusion:} The temporal traffic flow features $\text{TF}_{feat}$ are fused with the visual features of the lane through cross-modal interaction. The fused features are passed to a decoder to produce the final scene perception results. This design ensures that traffic flow features provide complementary road structure priors when visual cues are insufficient.

\subsection{Data Preparation}

\begin{figure}[h]
  \centering
  \begin{subfigure}{0.4\textwidth}
    \centering
    \includegraphics[width=\linewidth]{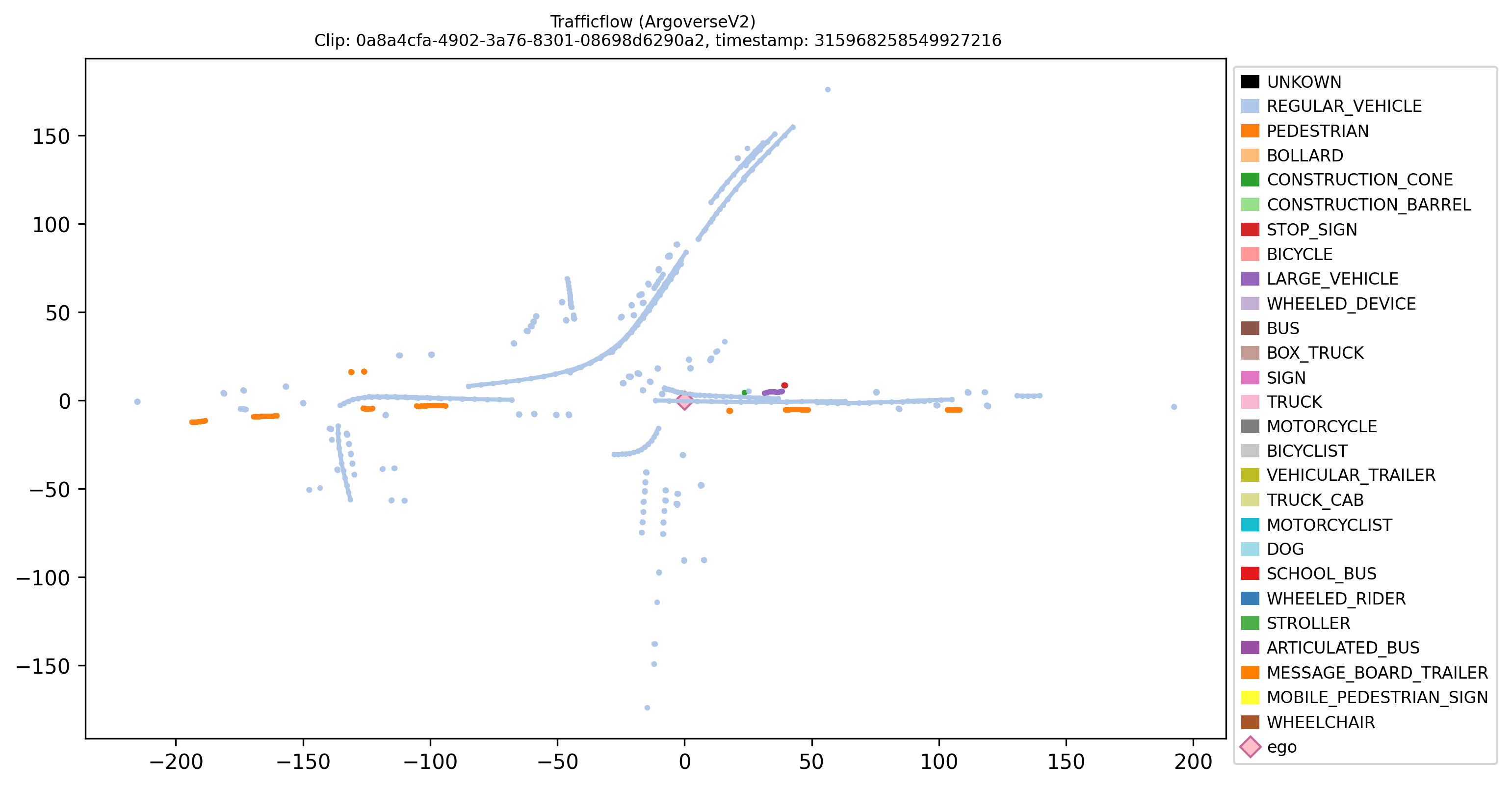}
    \caption{OpenLaneV2}
    \label{fig:toponet_score}
  \end{subfigure}
  
  \vspace{0.1cm}
  
  \begin{subfigure}{0.4\textwidth}
    \centering
    \includegraphics[width=\linewidth]{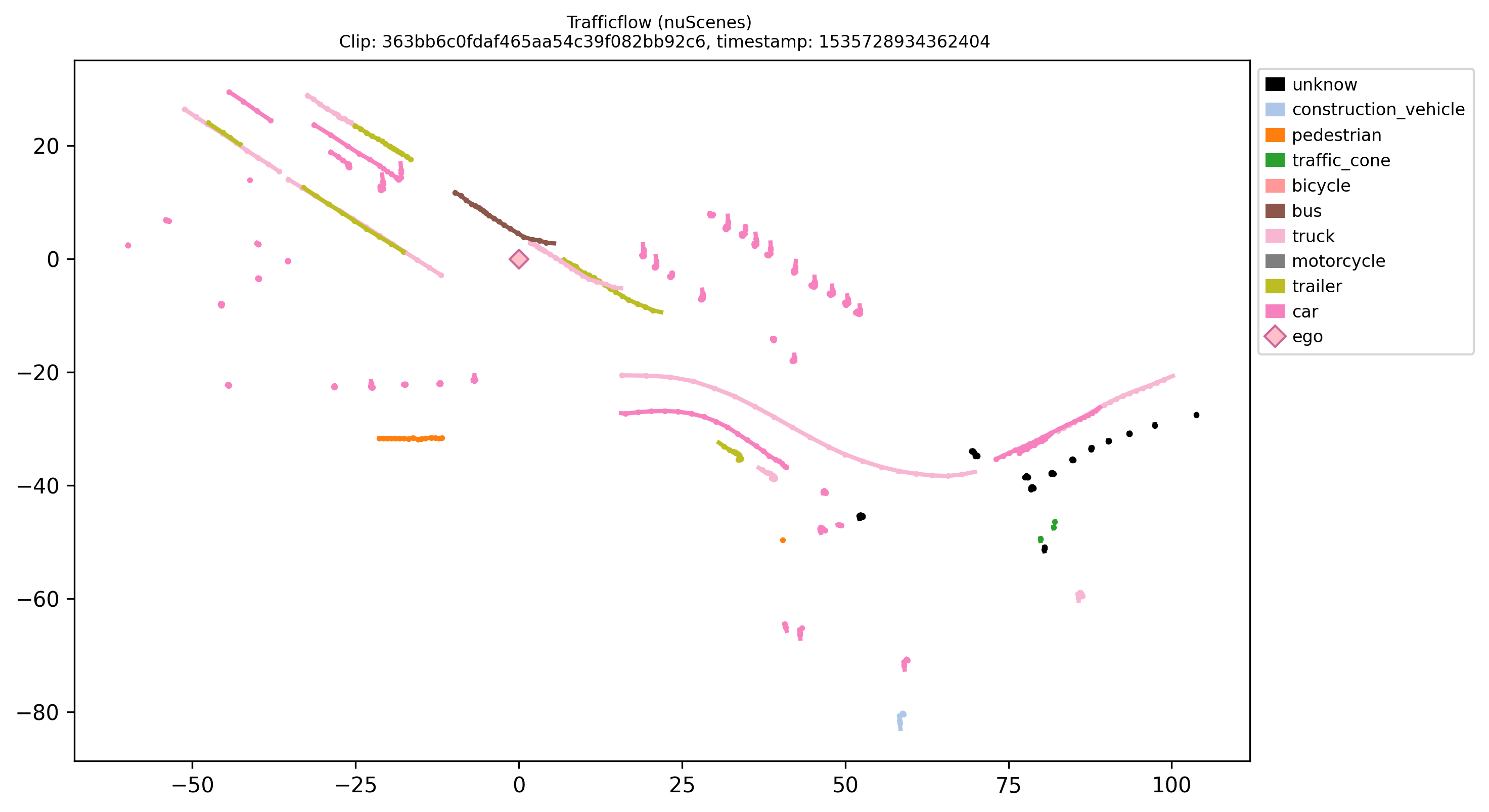}
    \caption{Nuscenes}
    \label{fig:lanesegnet_map}
  \end{subfigure}
  
  \caption{\textbf{Visualization of Traffic Flow in Datasets}}
  \label{fig:tf_datasets_vis}
  \vspace{-0.2cm}
\end{figure}

We process traffic flow data differently across datasets: for OpenLaneV2\cite{wang2023openlane}, we index Argoverse 2\cite{wilson2023argoverse} scene information via tokens to encode traffic flow, while for NuScenes\cite{caesar2020nuscenes}, we directly retrieve scene data through instance tokens. Specifically, given the current perception frame $f_i$ and its point cloud range $\mathcal{R}_p = [x_{\min}, x_{\max}] \times [y_{\min}, y_{\max}]$, we extract information from the scene, information—including tracking IDs, categories, trajectories, and occlusion states—from a historical frame sequence $\{f_{i-1}, f_{i-2}, ..., f_{i-n}\}$. The inter-frame pose transformation matrix:
\begin{equation}
    \mathbf{T}_{i-k}^i = \mathbf{Pose}_i^{-1} \cdot \mathbf{Pose}_{i-k}
\end{equation}
is applied to warp historical traffic flow information into the current frame's coordinate system, yielding aligned traffic flow representations $TF_i$. We visualize the traffic flow components of both datasets as shown in \cref{fig:tf_datasets_vis}, where it becomes evident that the spatial distribution of traffic flow allows us to infer structural information about the road network in the scene.

In particular, we preserve all valid traffic flow information without artificial smoothing, avoiding additional computational overhead while ensuring that the network receives the most real-time raw data. This design stems from our core insight: traffic flow should complement rather than replace visual perception, with its value particularly pronounced in complex scenarios where sensor cues are limited.

\subsection{Model Architecture}

Despite data preprocessing, dedicated network design remains essential for effectively learning and leveraging traffic flow priors. To this end, we propose a spatio-temporal dual-encoder architecture that enables efficient fusion between traffic flow features and lane features.

\paragraph{Temporal Domain Encoder}

We design a temporal domain encoder to model sequential characteristics of traffic flow while filtering invalid information. The encoder first truncates the information from the traffic flow to the perceptual region $\mathcal{R}_t$, retaining only elements relevant to the current scene. For historical $N$ frames, we set a validity threshold $tole_{pts}$: an instance is considered valid only if at least $tole_{pts}$ frames among consecutive $N$ frames contain valid traffic flow.

Taking into account the characteristics of the ego-motion and the spatial distribution patterns of the traffic flow, we employ an ego-centric weighting strategy. Specifically, traffic flow within a $60^\circ$ sector in front of the ego vehicle is assigned the highest weight, with weights progressively decreasing in other regions based on relative positions. Through this spatially sensitive mechanism, we extract the most representative $T_{top}$ traffic flow instances from the scene.

\begin{figure}[h]
  \centering
   \includegraphics[width=0.8\linewidth]{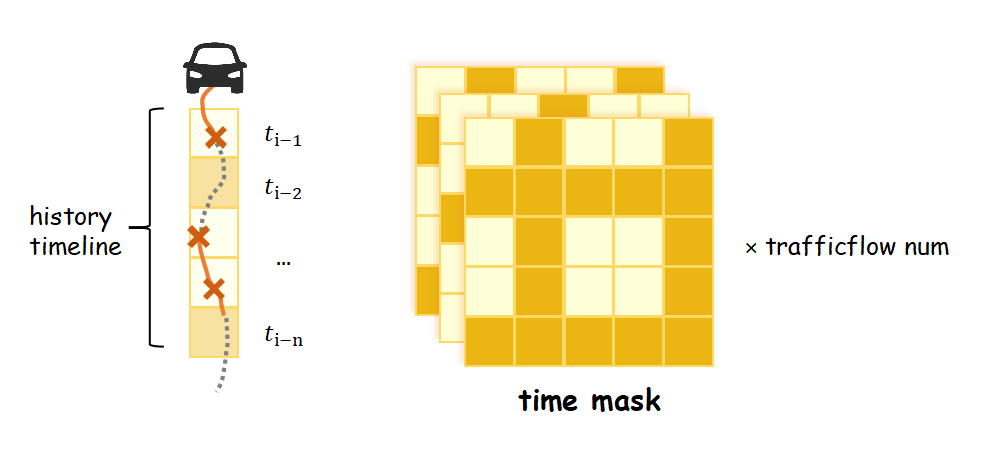}
   \caption{\textbf{Temporal Mask.} This mechanism focuses on the validity of historical frames within the temporal domain of traffic flow instances, effectively tackling real-world challenges such as frame drops and occlusions.}
   \label{fig:time-mask}
   \vspace{-0.2cm}
\end{figure}

In implementation, we set an upper bound $T_{\text{max}}$ for traffic flow instances per frame. When valid instances exceed $T_{\text{max}}$, we select based on weights; when insufficient, we pad to $T_{\text{max}}$. This ensures input consistency while avoiding redundancy.

Beyond generating refined traffic flow features $reTF_i$, the temporal encoder produces corresponding temporal masks, as shown in \cref{fig:time-mask}. These masks deactivate when temporal frames suffer from occlusion or information loss. This masking mechanism demonstrates strong robustness in real-world inference, effectively handling tracking interruptions caused by sensor communication failures and partial occlusions, thus ensuring stable performance in complex environments.

\paragraph{Spatial Domain Encoder}

\begin{figure}[htbp!]
  \centering
   \includegraphics[width=1\linewidth]{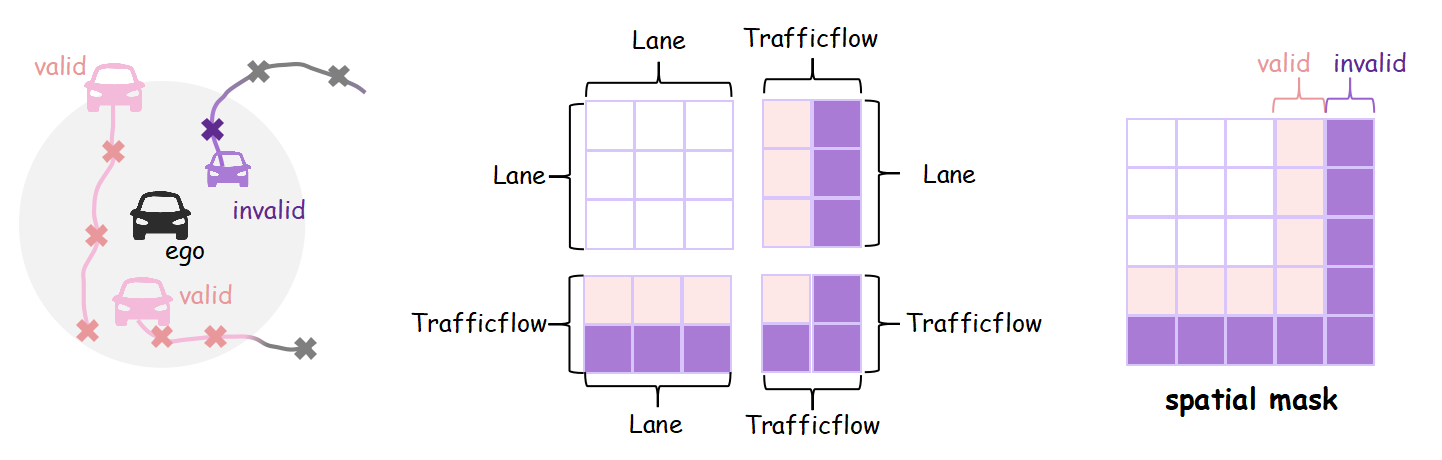}
   \caption{\textbf{Spatial Mask.} This mechanism identifies valid instances within current spatial partitions for both traffic flow and lane structures, effectively addressing challenges such as highly complex scenes and sparse instance distribution.}
   \label{fig:spatial-mask}
   \vspace{-0.2cm}
\end{figure}

The temporal encoder outputs enhanced features $\text{TF}_{Feat}$ and temporal masks, which are then processed by the spatial domain encoder. This encoder achieves structured fusion between traffic flow and lane features through a designed spatial masking mechanism, as shown in \cref{fig:spatial-mask}.

Let $L$ and $T$ denote the feature dimensions of lane and traffic flow representations, respectively. The spatial mask matrix $M_s \in \mathbb{R}^{(L+T)\times(L+T)}$ is divided into four logical regions:
\begin{itemize}
    \item $M_{L\rightarrow L}$: Intra-lane attention mask, reusing existing designs from the upstream network
    \item $M_{L\rightarrow T}$: Cross-modal attention mask from lanes to traffic flow
    \item $M_{T\rightarrow L}$: Cross-modal attention mask from traffic flow to lanes
    \item $M_{T\rightarrow T}$: Intra-traffic flow attention mask, generated based on instance validity
\end{itemize}

This masking mechanism explicitly models interaction patterns between modalities while ensuring focus on valid information. Specifically, we concatenate the spatial mask $M_s$, traffic flow features $\text{TF}_{Feat}$, and lane features $\text{L}_{Feat}$, then feed them into four fusion modules:
\begin{equation}
    F_{\text{fused}}^{(i)} = \mathcal{F}_{\text{fusion}}\left(\text{L}_{Feat}^{(i)}, \text{TF}_{Feat}^{(i)}; M_s^{(i)}\right), i=1,\dots,4
\end{equation}
where each module corresponds to a different partition of the mask matrix. The fused lane features are integrated with the original features through a paradigm-aware feature composition module. This process is formulated as:
\begin{equation}
    \text{L}_{Feat}^{'} = \mathcal{T}(F_{\text{fused}}^{(4)}) + I(Q_{\text{type}}) \odot \text{L}_{Feat}
\end{equation}
where $I(Q_{\text{type}})$ is an indicator that is 0 for instance-based queries and 1 for point-level or other structure-disrupting queries, and $\mathcal{T}$ denotes a feature transformation. This design provides the transformed fusion features $\mathcal{T}(F_{\text{fused}}^{(4)})$ as the primary input, while selectively adding the original $\text{L}_{Feat}$ as a residual only when needed to preserve structural information. The enhanced features $\text{L}_{Feat}^{'}$ are then passed to the decoder for final lane perception.


Notably, the $T\rightarrow T^{(1)}$ and $T\rightarrow L^{(2)}$ modules are optional components, while $L\rightarrow T^{(3)}$ and $L\rightarrow L^{(4)}$ modules are crucial for lane feature enhancement. Real-world deployment shows that activating only the latter two modules already brings significant performance gains with reduced computational overhead. This modular design offers deployment flexibility, adapting to various resource-constrained scenarios.

The proposed Traffic Flow-aware Module (TFM) enables efficient multi-modal feature fusion while preserving the original network architecture. This design offers architectural compatibility, deployment flexibility, and scene robustness, providing reliable environmental priors for lane detection that significantly enhance perception reliability in complex road scenarios.


\section{Experiments}

To validate the effectiveness of our approach, we conduct comprehensive evaluations on four mainstream open-source models and two public datasets. We demonstrate consistent performance gains across all baseline models, present qualitative analysis on challenging scenarios to showcase the value of traffic flow information, and perform ablation studies to verify each component's contribution.

\subsection{Implementation Details}

\paragraph{Metrics and Dataset} 
For thorough evaluation, we select four classical open-source models as baselines and strictly follow their original evaluation metrics and dataset configurations to ensure fair comparison:

\begin{itemize}
    \item \textbf{TopoNet}: Evaluated on OpenLaneV2 using perception metrics ($Det_t$, $Det_l$), topological reasoning metrics ($Top_{ll}$, $Top_{lt}$), and the overall metric $OLS$.
    \item \textbf{LaneSegNet}: Evaluated on OpenLaneV2 using the comprehensive metric $mAP$, lane perception metrics ($AP_{ls}$, $AP_{pred}$), and topological reasoning metric $TOP_{ll}$.
    \item \textbf{MapTR Series}: Evaluated on NuScenes using the overall $mAP$ and map element metrics ($AP_{\text{divider}}$, $AP_{\text{crossing}}$, $AP_{\text{boundary}}$).
\end{itemize}

Dataset selections align with the original papers of each model to maintain comparability.

\paragraph{Implementation Details}
All baseline models are trained and evaluated using 16 Alibaba Pingtouge PPUs for 60 epochs to ensure full convergence. Optimizer and learning rate hyperparameters follow official implementations. For the traffic flow module, we set the maximum number of traffic flow instances $N_t=30$ and historical trajectory frames $f_t=20$. Additional training details are provided in the appendix.

\subsection{Comparison}

To validate the generalization capability of TFM, we conduct systematic comparisons on four mainstream baselines. As shown in \crefrange{tab:lanesegnet_results}{tab:maptrv2_results}, TF-Lane achieves consistent performance improvements after integrating TFM, with the MapTR baseline attaining the highest gain of $+4.1\%$ \text{mAP} on the NuScenes dataset, demonstrating the significant prior value of traffic flow information.

\begin{figure*}[!htbp]
  \centering
  \begin{subfigure}{0.49\textwidth}
    \centering
    \includegraphics[width=\linewidth]{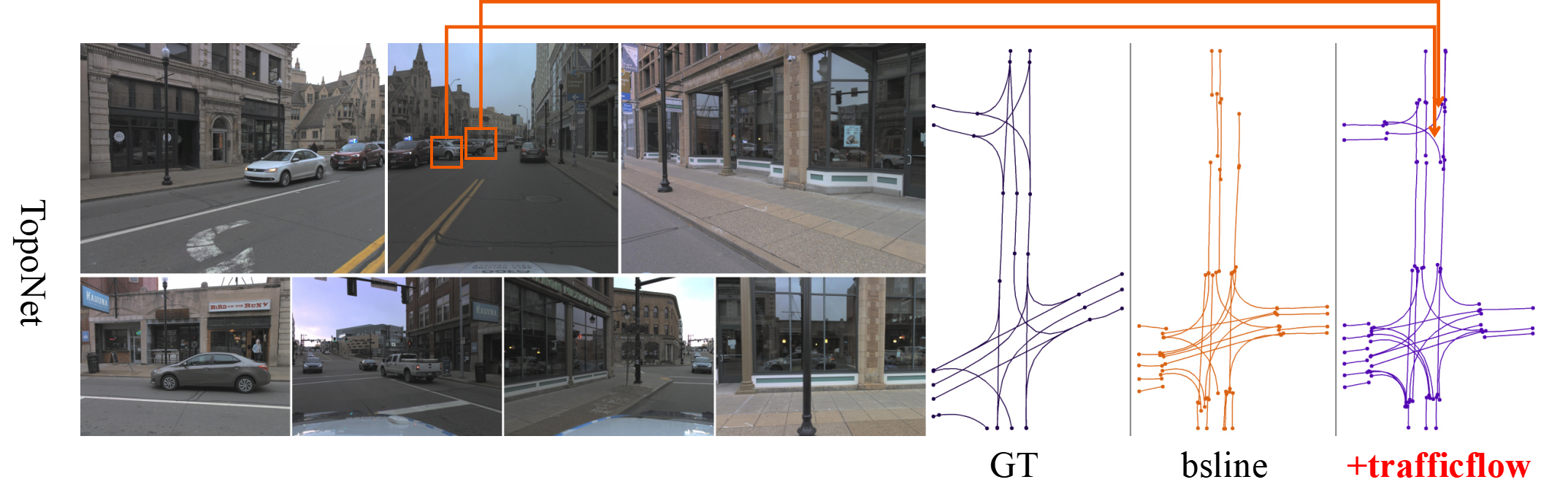}
    \caption{TopoNet visual comparison}
    \label{fig:toponet_infer}
  \end{subfigure}
  \hfill
  \begin{subfigure}{0.49\textwidth}
    \centering
    \includegraphics[width=\linewidth]{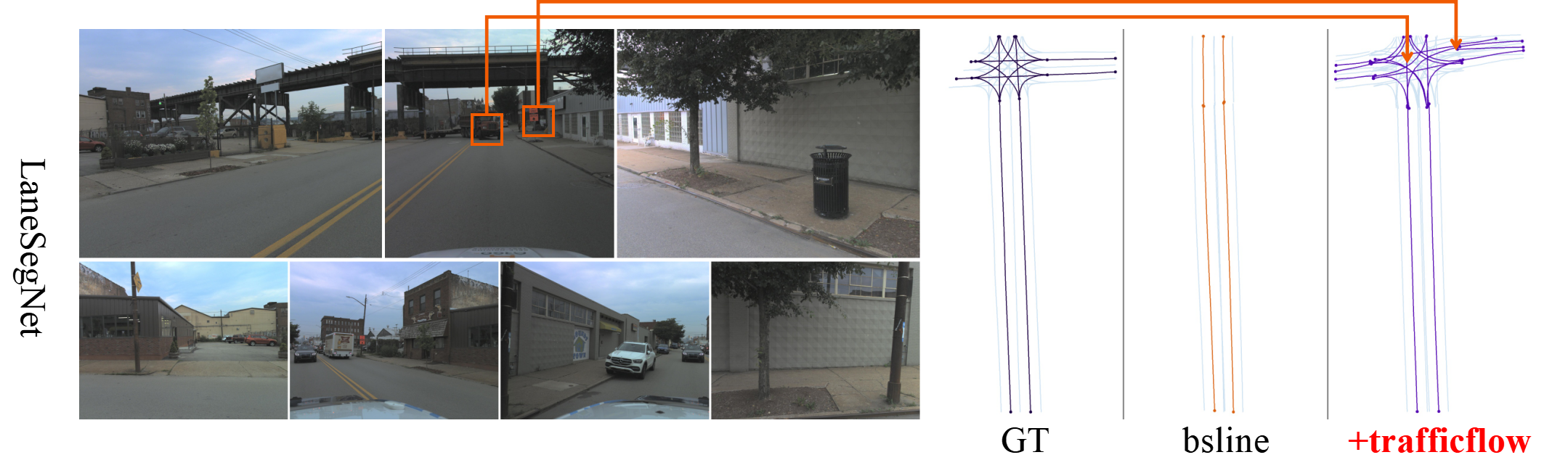}
    \caption{LaneSegNet visual comparison}
    \label{fig:lanesegnet_infer}
  \end{subfigure}
  
  \vspace{0.5cm}
  
  \begin{subfigure}{0.49\textwidth}
    \centering
    \includegraphics[width=\linewidth]{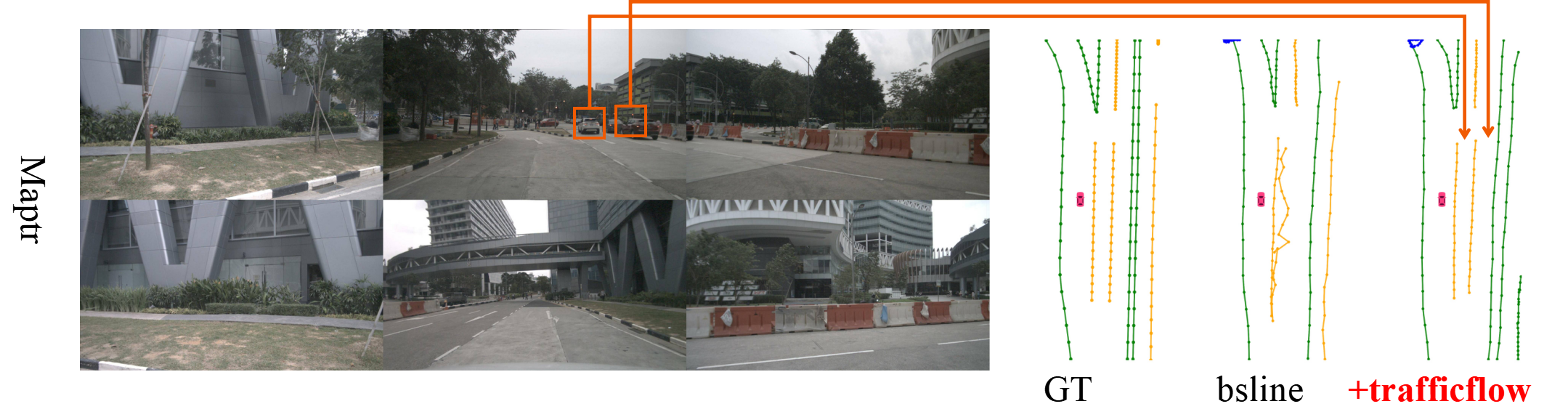}
    \caption{MapTR visual comparison}
    \label{fig:maptr_infer}
  \end{subfigure}
  \hfill
  \begin{subfigure}{0.49\textwidth}
    \centering
    \includegraphics[width=\linewidth]{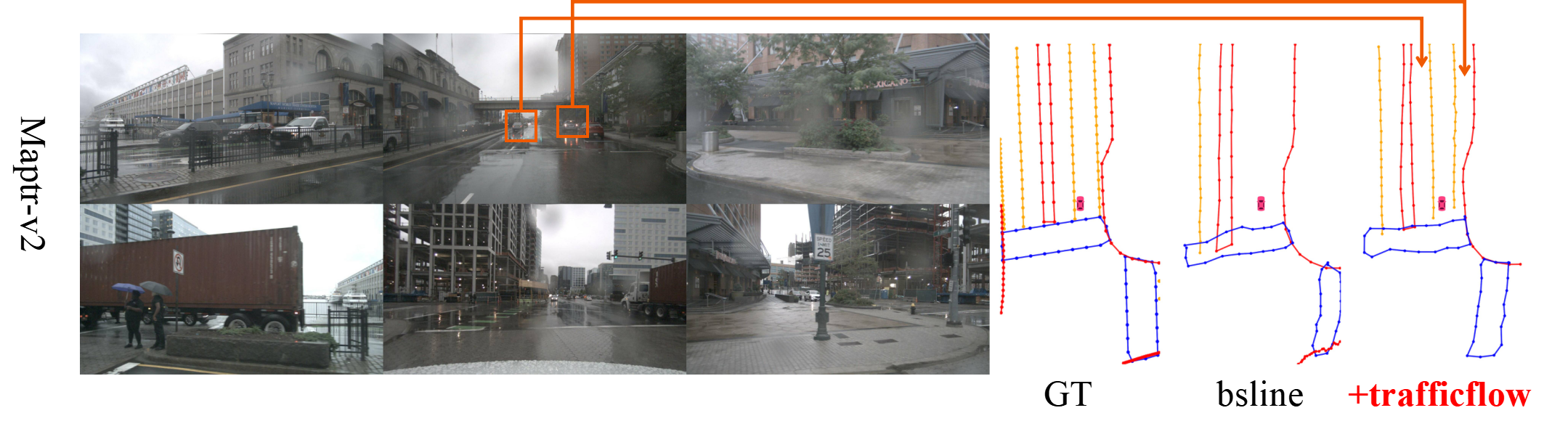}
    \caption{MapTRv2 visual comparison}
    \label{fig:maptrv2_infer}
  \end{subfigure}
  \caption{\textbf{Comparative Visualization.} A side-by-side comparison across four challenging scenarios: (a) T-junction diverges, (b) intersections with large vehicle occlusions, (c) distant traffic light intersections, and (d) rainy road conditions.}
  \label{fig:multi-model-infer}
  \vspace{-0.2cm}
\end{figure*}

To mitigate the impact of training randomness on evaluation, we employ a dual evaluation strategy: reporting both the final epoch (60) performance and the best results (*) during training. This design ensures the statistical reliability of performance comparisons and avoids random bias from single evaluations. Experiments show that TFM exhibits consistent improvement trends in all baseline models and evaluation metrics, verifying the universality and effectiveness of its architectural design.

\begin{table}[t]
\centering
\caption{\textbf{LaneSegNet Comparative Experiments.} Comparison results between traffic flow structure and the original baseline.}
\label{tab:lanesegnet_results}
\small
\begin{tabular}{@{}lc|cccccc@{}}
\toprule
Method & Epoch & \textbf{mAP} & $\mathrm{AP}_{ls}$ & $\mathrm{AP}_{pred}$ & $\mathrm{TOP}_{ll}$ \\
\midrule
Baseline & 60 & 32.99 & 32.75 & 33.24 & 27.02 \\
\textbf{Ours} & 60 & \textbf{35.02} & 34.90 & 35.15 & 27.83 \\
\addlinespace
Baseline & 50* & 34.65 & 34.04 & 35.26 & 26.76 \\
\textbf{Ours} & 46* & \textbf{35.66} & 34.52 & 36.79 & 27.50 \\
\bottomrule
\end{tabular}
\vspace{-0.1cm}
\end{table}

\begin{table}[t]
\centering
\caption{\textbf{TopoNet Comparative Experiments.} Comparison results between traffic flow structure and the original baseline.}
\label{tab:toponet_results}
\small
\begin{tabular}{@{}lc|cccccc@{}}
\toprule
Method & Epoch & \textbf{OLS} & $\mathrm{Det}_l$ & $\mathrm{Det}_t$ & $\mathrm{Top}_{ll}$ & $\mathrm{TOP}_{lt}$ \\
\midrule
Baseline & 60 & 41.12 & 31.08 & 44.22 & 15.44 & 24.89 \\
\textbf{Ours} & 60 & \textbf{42.29} & 31.27 & 47.20 & 15.97 & 25.74 \\
\addlinespace
Baseline & 55* & 41.22 & 31.06 & 44.62 & 15.45 & 24.91 \\
\textbf{Ours} & 56* & \textbf{42.39} & 31.22 & 47.44 & 15.97 & 25.96 \\
\bottomrule
\end{tabular}
\vspace{-0.2cm}
\end{table}

\begin{table}[!htbp]
\centering
\caption{\textbf{MapTR Comparative Experiments.} Comparison results between traffic flow structure and the original baseline.}
\label{tab:maptr_results}
\small
\begin{tabular}{@{}lc|ccccc@{}}
\toprule
Method & Epoch & \textbf{mAP} & $\mathrm{AP}_{div}$ & $\mathrm{AP}_{cross}$ & $\mathrm{AP}_{bd}$ \\
\midrule
Baseline & 60 & 51.51 & 53.92 & 47.76 & 52.86 \\
\textbf{Ours} & 60 & \textbf{55.61} & 59.28 & 50.05 & 57.52 \\
\addlinespace
Baseline & 56* & 51.63 & 54.12 & 47.86 & 52.91 \\
\textbf{Ours} & 50* & \textbf{55.80} & 58.75 & 51.10 & 57.57 \\
\bottomrule
\end{tabular}
\vspace{-0.1cm}
\end{table}

\begin{table}[!htbp]
\centering
\caption{\textbf{MapTRv2 Comparative Experiments.} Comparison results between traffic flow structure and the original baseline.}
\label{tab:maptrv2_results}
\small
\begin{tabular}{@{}lc|ccccc@{}}
\toprule
Method & Epoch & \textbf{mAP} & $\mathrm{AP}_{div}$ & $\mathrm{AP}_{cross}$ & $\mathrm{AP}_{bd}$ \\
\midrule
Baseline & 60 & 62.19 & 63.39 & 59.00 & 64.18 \\
\textbf{Ours} & 60 & \textbf{63.50} & 63.65 & 61.76 & 65.09 \\
\addlinespace
Baseline & 60* & 62.19 & 63.39 & 59.00 & 64.18 \\
\textbf{Ours} & 60* & \textbf{63.50} & 63.65 & 61.76 & 65.09 \\
\bottomrule
\end{tabular}
\vspace{-0.2cm}
\end{table}

  
  
  

\subsection{Ablation Study}

\paragraph{Architecture Components}

We conduct ablation studies on LaneSegNet to validate TFM's effectiveness, specifically comparing different cross-attention layer configurations: using all four spatial cross-attention layers, only the last two layers, and the effects of deepening layers and introducing normalized coordinates. As shown in \cref{tab:arch_ablation}, when the network depth is excessive (e.g., Abla1), the model overfits; while optimizing the depth (Abla2-Abla4) yields significant performance gains. Furthermore, encoding inputs as normalized coordinates aligned with BEV space (Abla5) further improves model robustness.

\begin{table}[!htbp]
\centering
\caption{\textbf{Architectural Ablation.} Evaluates TFM's structural design under fixed data settings ($N_t=30$, $tole_{pts}=5$).}
\label{tab:arch_ablation}
\scriptsize
\begin{tabular}{@{}lccc|cccc@{}}
\toprule
Setting & Pipe. & Dep. & Norm. & \textbf{mAP} & $\mathrm{AP}_{ls}$ & $\mathrm{AP}_{pred}$ & $\mathrm{TOP}_{ll}$ \\
\midrule
Baseline & - & - & - & 32.99 & 32.75 & 33.24 & 27.02 \\
Abla1 & All & 3 & F & 32.98 & 32.97 & 33.00 & 27.98 \\
Abla2 & All & 1 & F & 34.24 & 34.37 & 34.10 & 27.82 \\
Abla3 & LT/LL & 3 & F & 33.98 & 32.75 & 35.22 & 27.04 \\
Abla4 & LT/LL & 1 & F & 33.09 & 32.48 & 33.71 & 27.46 \\
Abla5 & \textbf{LT/LL} & \textbf{1} & \textbf{T} & \textbf{35.02} & 34.90 & 35.15 & 27.83 \\
\bottomrule
\end{tabular}
\end{table}

\begin{table}[!htbp]
\centering
\caption{\textbf{Data Ablation.} Evaluates TFM's data components under fixed architectural settings (Pipe.$=$ LT/LL, Dep.$=$1, Norm.$=$T)}
\label{tab:data_ablation}
\small
\begin{tabular}{@{}lcc|cccc@{}}
\toprule
\textbf{Setting} & N\_t & $tole_{pts}$ & \textbf{mAP} & $\mathrm{AP}_{ls}$ & $\mathrm{AP}_{pred}$ & $\mathrm{TOP}_{ll}$ \\
\midrule
Baseline & - & -  & 32.99 & 32.75 & 33.24 & 27.02 \\
Abla6 & 20 & 5 & 34.04 & 32.05 & 36.04 & 27.55 \\
Abla7 & \textbf{30} & \textbf{5} & \textbf{35.02} & 34.90 & 35.15 & 27.83 \\
Abla8 & 30 & 10 & 33.90 & 34.55 & 33.24 & 27.55 \\
\bottomrule
\end{tabular}
\end{table}

\begin{table}[!h]
\centering
\caption{\textbf{Generalization Ablation.} Evaluates the model's inference adaptability by constructing differentiated inference scenarios under fixed data and architectural parameters.}
\label{tab:infer_ablation}
\small
\begin{tabular}{@{}lcc|cccc@{}}
\toprule
Setting & $\text{train}_{N_t}$ & $\text{eval}_{N_t}$ & \textbf{mAP} & $\mathrm{AP}_{ls}$ & $\mathrm{AP}_{pred}$ & $\mathrm{TOP}_{ll}$ \\
\midrule
Baseline & - & - & 32.99 & 32.75 & 33.24 & 27.02 \\
Abla9 & 30 & \textbf{30} & \textbf{35.02} & 34.90 & 35.15 & 27.83 \\
Abla10 & 30 & \textbf{0} & \textbf{34.32} & 33.54 & 35.10 & 27.82 \\
\bottomrule
\end{tabular}
\end{table}

\paragraph{Data Processing Strategies}

Beyond architectural design, we analyze data parsing strategies. Here, $N_t$ denotes the number of traffic flow instances introduced during training (with padding when insufficient), and $tole_{\text{pts}}$ defines the valid frame threshold for each traffic flow instance, beyond which samples are considered valid. As shown in \cref{tab:data_ablation}, increasing traffic flow quantity (e.g., Abla7) brings greater performance gains. The threshold for valid instances can be dynamically adjusted according to the object detection module of the autonomous driving system to balance data quality and model generalization.

\paragraph{Generalization Capability}

To further investigate TFM's generalization, we design a partial-modality inference testing: training with traffic flow data while inferring without traffic flow data. As shown in \cref{tab:infer_ablation}, TFM still achieves stable improvements over the baseline even without traffic flow information during inference, indicating that traffic flow features can serve as auxiliary supervision signals to enhance representation learning. This result verifies that TFM not only suits explicit end-to-end models but also improves performance in implicit networks (without traffic flow inputs during inference).

\subsection{Qualitative Visualization}

  
  

To validate TFM's generalization capability, we conduct visual comparisons of inference results across four benchmark models. Compared to baseline methods, models integrated with TFM demonstrate more robust perception performance across all test scenarios. We specifically select four challenging scenarios for inference and visualization (as shown in \cref{fig:multi-model-infer}): intersections with large vehicle occlusions (LaneSegNet), T-junction diverges (TopoNet), distant traffic light intersections (MapTR), and rainy road conditions (MapTRv2). Visualization results indicate that by incorporating highly real-time traffic flow priors, TFM significantly enhances adaptability to complex scenarios like occlusions and adverse weather conditions, producing perception outputs that better align with actual road structures and validating the effectiveness of our module design.

\begin{figure}[t]
  \centering
  \begin{minipage}{\textwidth}
    \begin{subfigure}{0.23\textwidth}
      \centering
      \includegraphics[width=\linewidth]{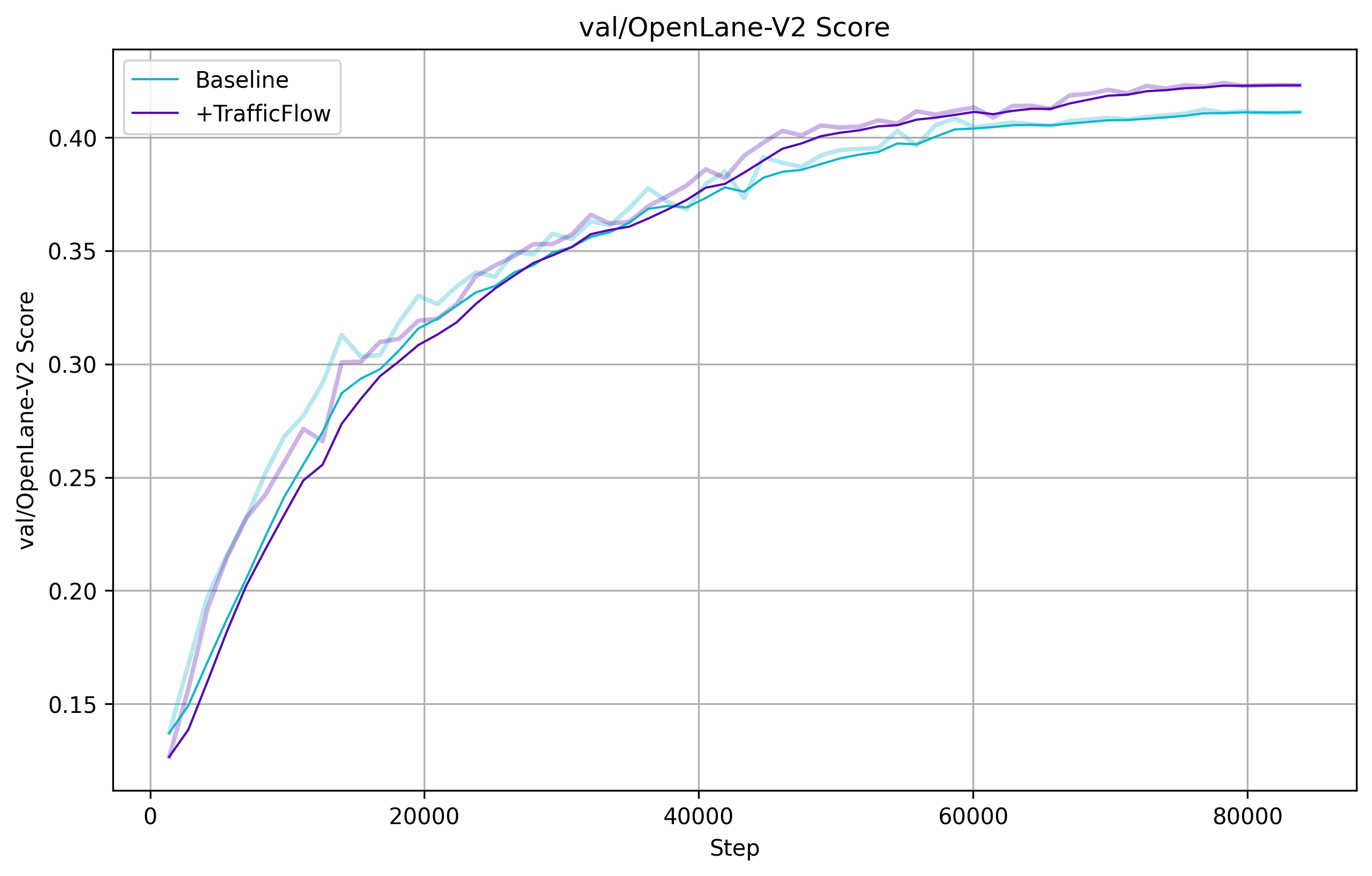}
      \caption{TopoNet's OLS}
      \label{fig:toponet_score}
    \end{subfigure}
    \begin{subfigure}{0.23\textwidth}
      \centering
      \includegraphics[width=\linewidth]{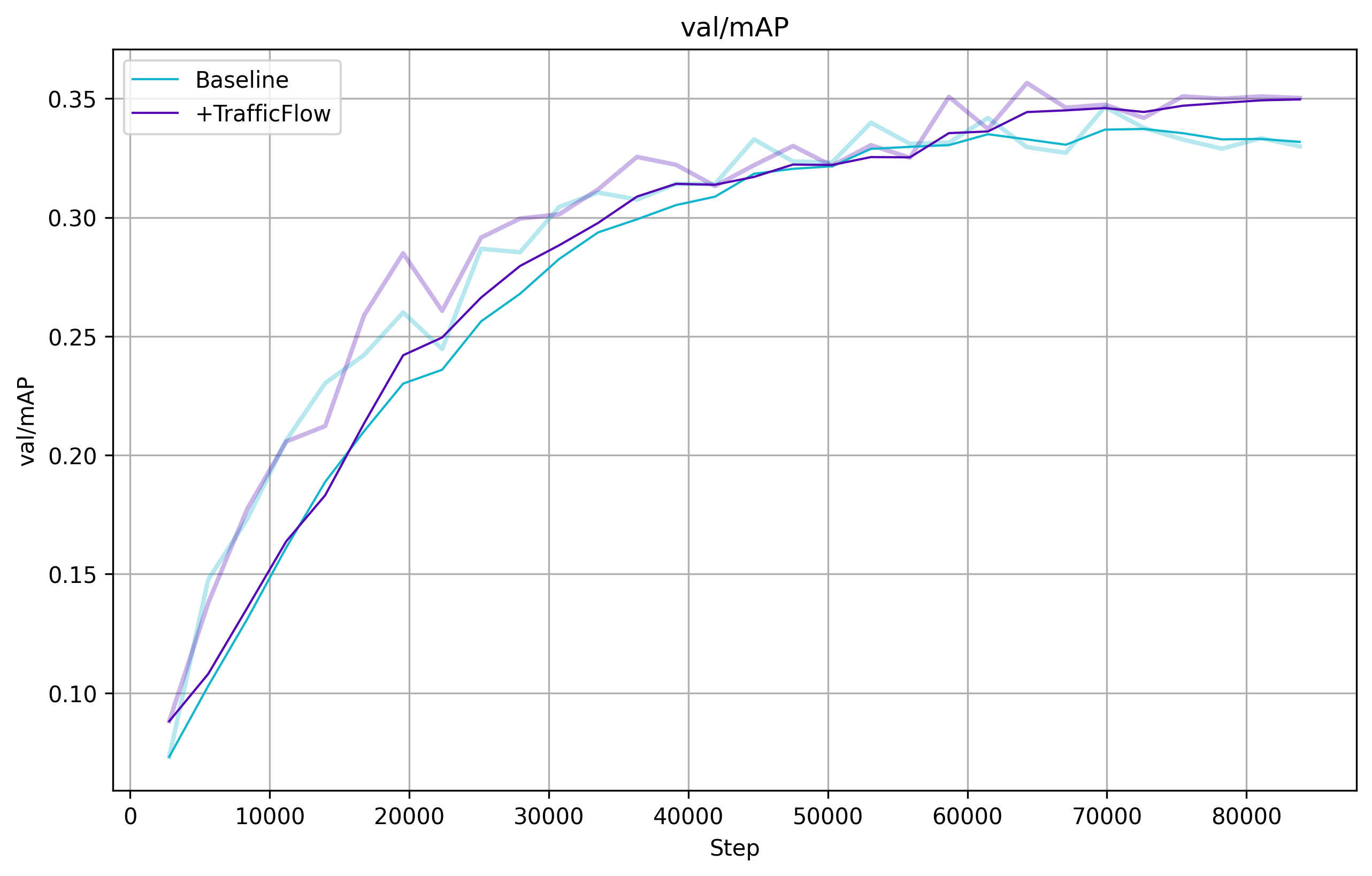}
      \caption{LaneSegNet's mAP}
      \label{fig:lanesegnet_map}
    \end{subfigure}
  \end{minipage}
  
  \vspace{0.2cm}
  
  \begin{minipage}{\textwidth}
    \begin{subfigure}{0.23\textwidth}
      \centering
      \includegraphics[width=\linewidth]{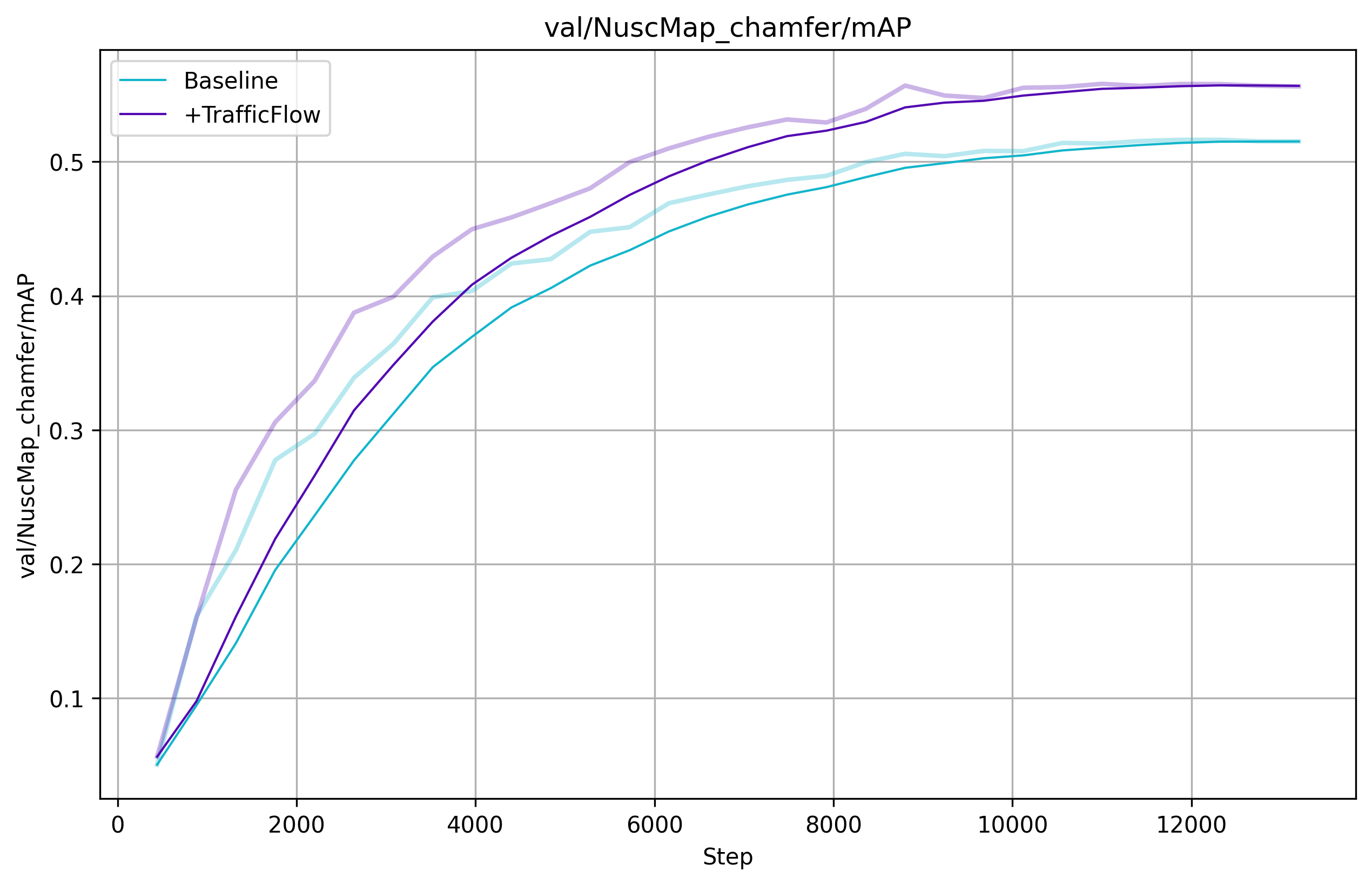}
      \caption{MapTR's mAP}
      \label{fig:maptrv1_map}
    \end{subfigure}
    \begin{subfigure}{0.23\textwidth}
      \centering
      \includegraphics[width=\linewidth]{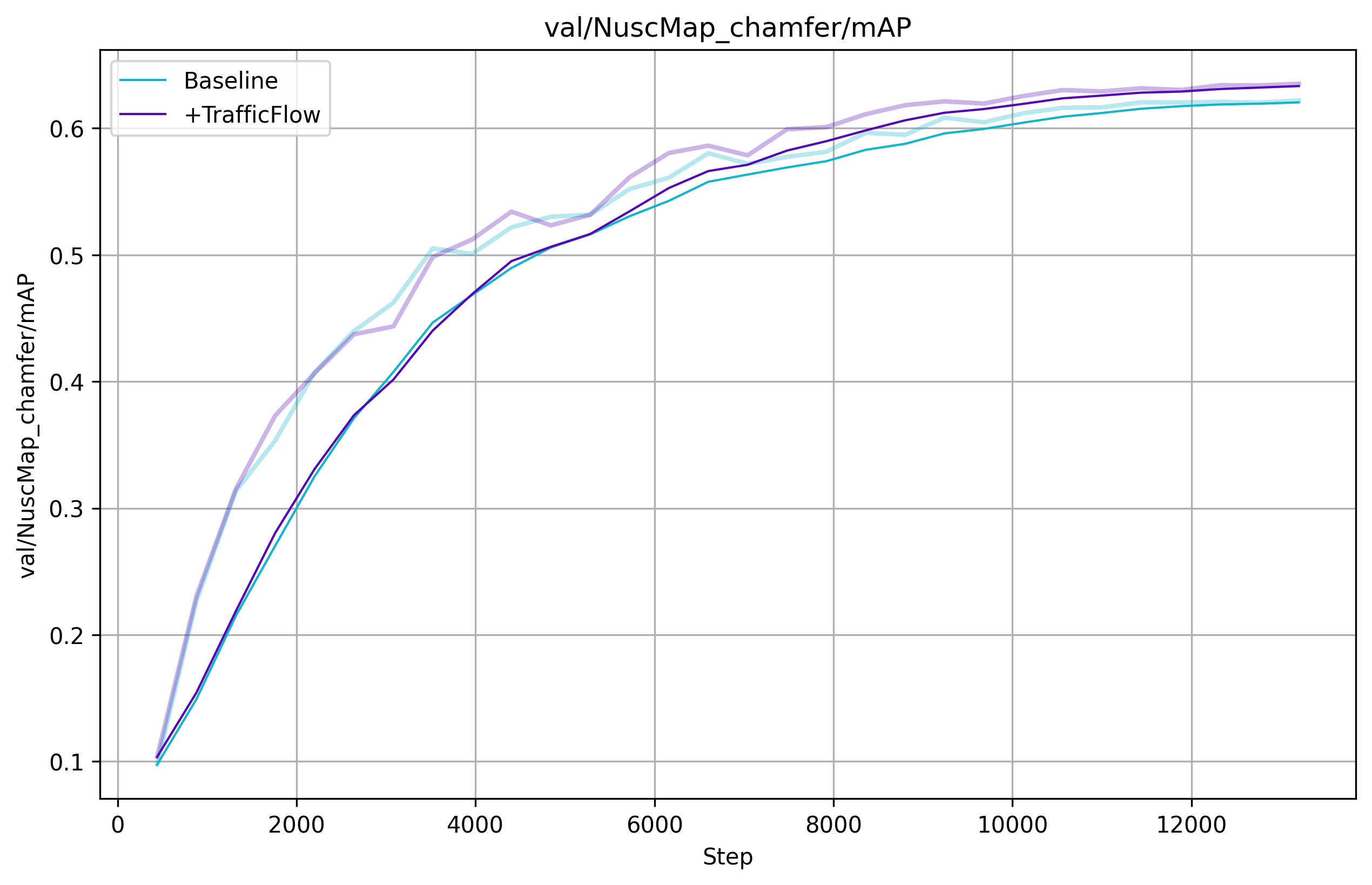}
      \caption{MapTRv2's mAP}
      \label{fig:maptrv2_map}
    \end{subfigure}
  \end{minipage}
  
  \caption{\textbf{Validation Set Metrics Comparison.} Performance curves of four baseline models across key evaluation metrics.}
  \label{fig:model_comparison}
\end{figure}

\section{Conclusion}

This paper presents a traffic flow prior-based enhancement method for lane detection. By introducing zero-subscription-cost, highly real-time traffic flow features, the proposed TFM module is compatible with a wide range of existing detection frameworks, effectively addressing perception degradation in complex scenarios such as occlusions and lane missing. Extensive experiments on two public datasets and four mainstream models demonstrate significant performance improvements under standard evaluation metrics (e.g., up to +4.1\% mAP on NuScenes), with ablation studies confirming the robustness of both module architecture and data strategies.

Our work originates from practical requirements in real-world autonomous driving systems and has been successfully deployed in industrial applications with closed-loop validation, demonstrating a complete pathway from industrial practice to academic contribution.
{
    \small
    \bibliographystyle{ieeenat_fullname}
    \bibliography{main}
}


\end{document}